\documentclass[10pt,twocolumn,letterpaper]{article}

\usepackage[pagenumbers]{iccv}

\usepackage{customization}

\definecolor{iccvblue}{rgb}{0.21,0.49,0.74}
\usepackage[pagebackref,breaklinks,colorlinks,allcolors=iccvblue]{hyperref}

\def\paperID{7}
\def\confName{ICCV}
\def\confYear{2025}

\title{Joint Training of Image Generator and Detector for Road Defect Detection}

\author{Kuan-Chuan Peng\thanks{Acknowledgement: Ramy Mounir (Univ. of South Florida; ramy@usf.edu) made some technical contributions to this work as an intern at MERL.}\\
Mitsubishi Electric Research Laboratories (MERL), Cambridge, MA, USA\\
{\tt\small kpeng@merl.com}
}

\begin{document}
\maketitle
\begin{abstract}
Road defect detection is important for road authorities to reduce the vehicle damage caused by road defects. Considering the practical scenarios where the defect detectors are typically deployed on edge devices with limited memory and computational resource, we aim at performing road defect detection without using ensemble-based methods or test-time augmentation (TTA). To this end, we propose to \textbf{J}ointly \textbf{T}rain the image \textbf{G}enerator and \textbf{D}etector for road defect detection (dubbed as \ours). We design the dual discriminators for the generative model to enforce both the synthesized defect patches and overall images to look plausible. The synthesized image quality is improved by our proposed CLIP-based Fréchet Inception Distance loss. The generative model in \ours is trained jointly with the detector to encourage the generative model to synthesize harder examples for the detector. Since harder synthesized images of better quality caused by the aforesaid design are used in the data augmentation, \ours outperforms the state-of-the-art method in the RDD2022 road defect detection benchmark across various countries under the condition of no ensemble and TTA. \ours only uses less than 20\% of the number of parameters compared with the competing baseline, which makes it more suitable for deployment on edge devices in practice.
\end{abstract}    
\section{Introduction}
\label{sec:intro}

Potholes and road defects are menacing threats to the safety of the road users, including the drivers, bicyclists, and pedestrians. In the period spanning 2018 to 2020, these road imperfections tragically claimed the lives of more than 5000 individuals in India~\cite{motive1}. In addition, out of every ten drivers, one experienced vehicle damage substantial to necessitate repairs following encounters with potholes, which resulted in an average repair cost nearing \$600 per incident, culminating in a staggering total of \$26.5 billion in damages for the year 2021 alone~\cite{motive2}. The aforesaid statistics makes detecting road defects an indispensable task for the road authorities and motivates road damage detection challenges in computer vision, \eg, the CRDDC 2022 challenge~\cite{crddc}.

The state-of-the-art (SOTA) defect detection methods and the top performing methods of the CRDDC 2022 challenge often require ensemble from multiple models~\cite{sota_baseline,sota_wang,sota_okran,sota_bhavsar,sota_jeong} or performing test-time augmentation (TTA)~\cite{sota_baseline,sota_wang}. Given the real-world scenarios where the defect detectors are commonly deployed on edge devices characterized by constrained memory and computational resources (\eg, no GPU)~\cite{iskd}, using ensemble-based method or performing TTA is impractical due to their computational overhead at the testing time, and the trained model ideally needs to be lightweight. Therefore, we challenge ourselves the following problem: \textit{Without using any ensemble-based techniques or any form of test-time augmentation, what can practitioners do to improve the performance of road defect detection using fewer number of model parameters at the testing time}?

One common technique to improve the performance of the road defect detectors is to perform data augmentation~\cite{sota_saha,sota_baseline,sota_okran,sota_jeong}. Nevertheless, these methods typically only use limited predefined low-level augmentation techniques such as color jitter, horizontal flip, \etc. Some methods use generative models to synthesize realistic and diverse examples~\cite{maeda,zhong}. However, the models trained by these methods are only trained to optimize the synthesized defect patch to be plausible, but they are not trained to make the entire image plausible as well, \eg,~\cite{maeda,zhong} use the Poisson blending to blend the synthesized defect patch into the background image post hoc. \cite{maeda} even identifies that such blending method can be the cause of their suboptimal performance. To resolve the issue in these works and optimize the entire image to be plausible at the training time, we propose to use the dual discriminator architecture in the generative model such that the realism of both the synthesized defect patch and the overall image (with the synthesized defect in it) is enforced.

Another issue of using the generative model which is totally overlooked by the road defect detection literature is that the generative model is typically trained separately from the detector in advance. Due to such practice, the generative model in the prior works is unaware of the existence of the detector and its objective (\ie, detect the defects). The only objective of the generative model in the prior works is to synthesize plausible images which have no guarantee to be beneficial to the detector's performance, and hopefully such implicit objective can enhance the detector's performance. Instead of following the suboptimal practice of the prior works, we propose to train the generative model and detector \textbf{jointly} such that enhancing the detector's performance becomes part of the major objective of the generative model, which we believe is a novel and strong departure from the prior works. Specifically, we use the concept of hard example mining and propose a novel loss term for the generative model to encourage the generative model to synthesize more challenging examples that the detector fails to detect.

Ideally, the images synthesized by the generative models should be of the same distribution as the real training data for the purpose of data augmentation. Therefore, we expect that the quality of the synthesized images and that of the real training images should be as close as possible. To further improve the synthesized image quality, we propose to minimize the Fréchet inception distance (FID)~\cite{fid} as part of the optimization when training the generative model. Although FID has become a standardized evaluation metric to assess image quality, surprisingly, we are unaware of any defect detection works which explicitly optimize FID during training to improve the synthesized image quality for the data augmentation purpose.

To the best of our knowledge, we are the first in road defect detection to jointly train the generative model and detector, use dual discriminators, and explicitly optimize the FID as part of the objectives when training a generative model for data augmentation. We make the following contributions:
\begin{enumerate}
    \item We propose to \textbf{J}ointly \textbf{T}rain an image \textbf{G}enerator and \textbf{D}etector for the task of road defect detection (dubbed as \ours). Under the condition of no ensemble-based techniques or test-time augmentation, \ours outperforms the state-of-the-art (SOTA) defect detection method on the RDD2022 dataset~\cite{rdd} in most cases of country-wise and overall performance across six countries.
    \item For road defect detection, we propose the hard example synthesis loss for jointly training the generative model and detector, the CLIP-based Fréchet Inception Distance loss, and the dual discriminator architecture to improve the image synthesis quality of the generative model.
    \item Outperforming the SOTA baseline, \ours only uses less than 20\% of the test-time model parameters compared with the SOTA baseline, which supports that \ours is a more suitable defect detection method to be deployed on the edge devices in the real-world applications where the memory and computational resources are limited.
    \item We provide both the qualitative and quantitative ablation study to support that jointly training the generative model and detector, and enhancing the synthesized image quality using our proposed techniques can improve the defect detection performance.
\end{enumerate}

\section{Proposed Method -- \ours}
\label{subsec:method}

\begin{figure}
\centering
\includegraphics[width=\linewidth]{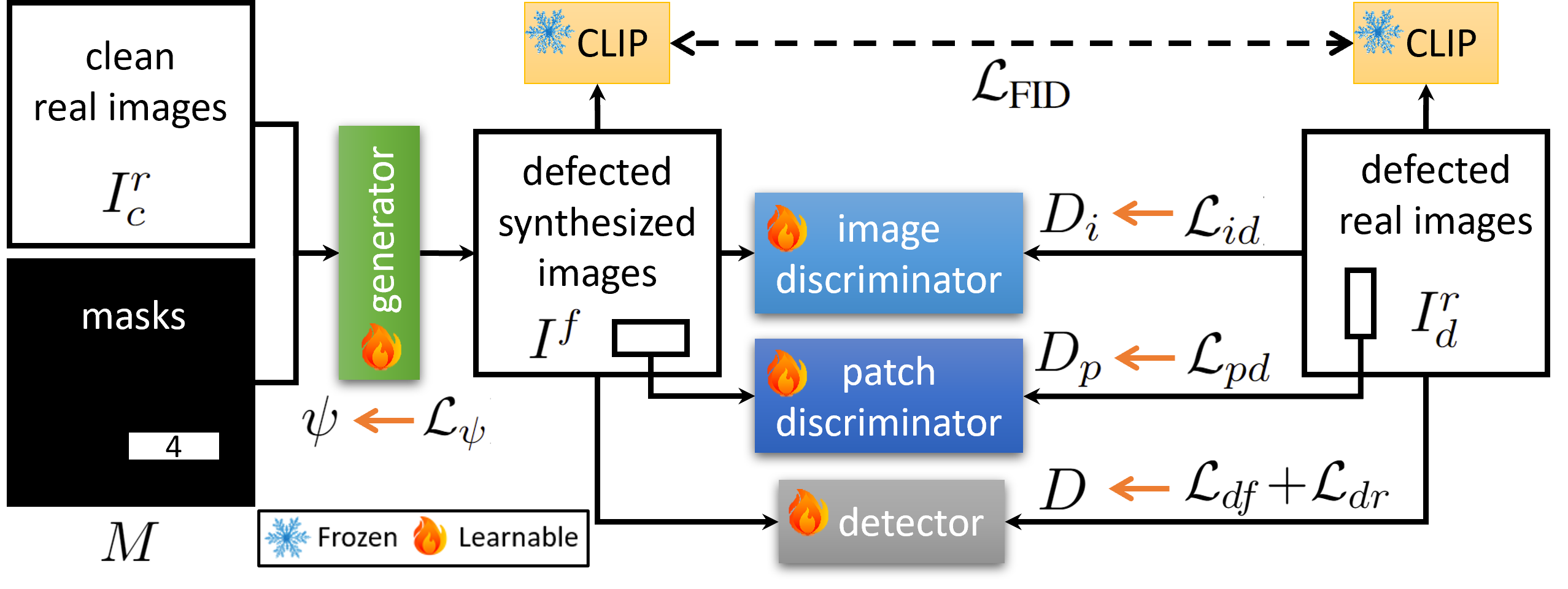}
\caption{The overview of our method \ours, where we propose to use the dual discriminators ($D_i$ and $D_p$) and the Fréchet inception distance loss $\cL_\text{FID}$ computed using the CLIP features (frozen) to improve the quality of the images synthesized by the generator $\psi$ for data augmentation when training the defect detector $D$. We train $\psi$ and $D$ \textbf{jointly} such that $\psi$ can learn to synthesize harder examples based on $D$'s detection results, which makes it possible for $\psi$ to update itself based on $D$'s objective (defect detection). The orange arrows point to the locations where each loss is enforced on. See Sec.~\ref{subsec:method} for details.
}
\label{ours}
\vspace{-.5em}
\end{figure}

Our proposed method \ours is illustrated in Fig.~\ref{ours}, where we propose to use the dual discriminators (\ie, image discriminator $D_i$ and patch discriminator $D_p$) and the Fréchet inception distance loss $\cL_\text{FID}$ with the CLIP~\cite{clip} features to improve the quality of the images synthesized by the image generator $\psi$ for the data augmentation purpose. Furthermore, $\psi$ is trained jointly with the detector $D$ such that the detection result from $D$ serves as the feedback for $\psi$ to encourage $\psi$ to generate harder examples for $D$ to detect.
In Fig.~\ref{ours}, we only show the part of training the generative model jointly with the detector for clarity. Given a set of clean (defect-free) real images $I^r_c$ and their masks $M$ indicating where and what types of the defects we would like to add to $I^r_c$ (the number in the box of $M$ in Fig.~\ref{ours} illustrates the defect type ID), the goal of the image generator $\psi$ is to synthesize the fake images $I^f$ with the defect types and locations specified in $M$ such that the defects seamlessly blend into the same background of $I^r_c$. Different from the prior works~\cite{maeda,zhong} which only employ one discriminator ($D_p$ in Fig.~\ref{ours}) to enforce that the synthesized defect patches are indistinguishable from the defect patches of the real training images, we add an additional image discriminator $D_i$ to enforce that the entire $I^f$ are also indistinguishable from the entire real defected images $I^r_d$. Such dual discriminator design encourages not only the synthesized defect patches but also the overall synthesized defected images to be plausible. Formally, the objective function of $D_i$ (dubbed as the image discriminator loss $\cL_{id}$) can be written as:
\begin{equation}
\cL_{id}=-\EE_{r_i\sim R_i}[\text{log}D_i(r_i)]-\EE_{f_i\sim F_i}[\text{log}(1-D_i(f_i))],
\end{equation}
where $R_i$ is the data distribution of $I^r_d$, and $F_i$ is the data distribution of $I^f$. The objective function of $D_p$ (dubbed as the patch discriminator loss $\cL_{pd}$) can be written as:
\begin{equation}
\cL_{pd}=-\EE_{r_p\sim R_p}[\text{log}D_p(r_p)]-\EE_{f_p\sim F_p}[\text{log}(1-D_p(f_p))],
\end{equation}
where $R_p$ and $F_p$ denote the distributions of real defect patches in $I^r_d$ and synthesized defect patches in $I^f$, respectively.

To further improve the synthesized image quality, we propose to add the Fréchet inception distance loss $\cL_\text{FID}$ as an additional objective of $\psi$. The goal of $\cL_\text{FID}$ is to encourage the statistics of $I^f$ and $I^r_d$ to be the same. Inspired by the recent finding~\cite{fid_clip} that the FID evaluated using the ImageNet~\cite{imagenet} features does not necessarily reflect the image quality, we modify the existing ImageNet-based FID loss implementation~\cite{fid_loss} such that the FID is evaluated using the CLIP~\cite{clip} features, as suggested by~\cite{fid_clip}. To be clear, we are the first in road defect detection which utilizes the foundation model (FM) features (\eg, CLIP) to improve data augmentation quality, and hence the final performance. Most of the prior works which use the concept similar to FID did not even consider leveraging the large FMs nowadays but still use the ImageNet based features which are proven to fail to reflect image quality~\cite{fid_clip}, which clearly separates \ours from existing works.

In addition to the aforesaid design, the most distinguishing characteristic of \ours is that $\psi$ is \textbf{jointly} trained with the detector $D$ such that $\psi$ can adjust the synthesized image quality according to the feedback given by $D$. This design improves over the traditional data augmentation pipeline where the image generator and detector are trained \textbf{separately} (\ie, an image generator is trained first, and a detector is trained later using the images synthesized by the already trained and frozen image generator). One clear drawback of the traditional data augmentation pipeline is that the image generator is totally unaware of the objective of the task of interest (\eg, road defect detection) because the objective of the image generator in the traditional data augmentation pipeline typically only encourages the image generator to synthesize plausible images (which is not directly related to our task of interest), instead of encouraging the image generator to \textit{synthesize the images which will be beneficial to the objective of the task of interest}. To the best of our knowledge, we are the first in road defect detection to propose jointly training the generative model and detector to optimize the detector's performance, which is a strong departure from prior works in road defect detection.

To realize the aforementioned idea and let $\psi$ and $D$ be trained jointly, we feed $I^f$ and $I^r_d$ as the input of $D$ and supervise $D$ with the detection losses $\cL_{df}$ and $\cL_{dr}$, respectively, where $\cL_{df}$ and $\cL_{dr}$ are the same detection loss as that used in the Faster Swin~\cite{sota_baseline} except that they correspond to different inputs ($I^f$ and $I^r_d$). Since we use $\psi$ to synthesize $I^f$ on the fly during training, we know exactly where the synthesized defects are placed in $I^f$, which provides the supervision signal for $\cL_{df}$. The supervision signal for $\cL_{dr}$ is directly from the ground truth of the training data. Given $\cL_{df}$ and $\cL_{dr}$, we propose the following hard example synthesis loss $\cL_h$ as an additional loss term to supervise $\psi$:
\begin{equation}
\cL_h = -\cL_{df}-\cL_{dr}.
\end{equation}
$\cL_h$ encourages $\psi$ to synthesize harder examples where $D$ fails to detect the defects, which is similar to the spirit of hard example mining. Once the synthesized hard examples are added to the training set and we retrain $D$, the performance of the retrained $D$ is expected to improve because its ability to detect harder examples is enhanced.

Formally, the final objective function of $\psi$ (dubbed as $\cL_\psi$) can be written as: $\cL_\psi=\cL_g + w_h\cL_h + w_\text{FID}\cL_\text{FID}$,
where $w_h$ and $w_\text{FID}$ are the weights of 
$\cL_h$ and $\cL_\text{FID}$, respectively, and
\begin{equation}
\cL_g=\EE_{f_i\sim F_i}[\text{log}(1-D_i(f_i))]+\EE_{f_p\sim F_p}[\text{log}(1-D_p(f_p))].
\end{equation}
After training $\psi$, we use it to synthesize defected images to augment the training set for $D$. $D$ is then retrained with the union of the original real training set and the set of the synthesized training images using the same detection loss as that used in the Faster Swin~\cite{sota_baseline} such that the retrained $D$ can perform well on not only the original training images but also more challenging synthesized images, which we expect to enhance $D$'s generalization ability on the unknown testing set.

\section{Experimental setup}

\begin{table*}
\centering
\resizebox{\textwidth}{!}{%
\begin{tabular}{ccccccccc}
\toprule
experiment ID &method &\# parameters (M)$\downarrow$ &India &Japan &Norway &USA &mean (left 4 countries) &overall (6 countries)\\
\midrule
$E_1$ &Faster Swin &253 &39.82 &60.57 &34.67 &\textbf{66.71} &50.44 &59.20\\
$E_5$ &\ours &\textbf{49} &\textbf{39.86} &\textbf{61.88} &\textbf{47.28} &66.24 &\textbf{53.82} &\textbf{63.13}\\
\bottomrule
\end{tabular}}
\caption{
\ours outperforms the top-performing method Faster Swin~\cite{sota_baseline} in average F1 scores across 4 listed countries and overall across 6 (including Czech Republic and China) countries in the CRDDC 2022 challenge~\cite{crddc}, while using $<$20\% of its parameters. All F1 values (\%) are from the RDD2022 test set, evaluated by the challenge server.
}
\label{result}
\vspace{-1.8em}
\end{table*}

We experiment on the RDD2022 road damage detection dataset~\cite{rdd}, one of the largest public datasets for road defect detection, associated with the CRDDC 2022 challenge~\cite{crddc}. Following the evaluation protocol of the CRDDC 2022 challenge~\cite{crddc}, we present the statistics of the RDD2022 dataset and the details of the evaluation protocol in the Apendix.

We use the InternImage-T~\cite{ii} as the backbone of the detector because it is one of the SOTA lightweight backbones across general object detection benchmarks such as PASCAL VOC~\cite{pascal} and COCO~\cite{coco}. We adapt the CycleGAN~\cite{cyclegan} as the backbone of the generative model such that it can take both a defect-free image and a mask indicating where to add synthesized defects as input and implement dual discriminators in \ours using the default discriminator architecture in CycleGAN. We use the road images in the RDD2022 training set which do not have any defect annotation as the defect-free images to train the generator. We perform drivable area detection as the preprocessing step to ensure that the synthesized defects are placed within the road areas (details are in the Appendix). We train the modified CycleGAN (generator) from scratch with dual discriminators and $\cL_\text{FID}$ until convergence. Once the generator is trained, we use it to synthesize road images with defects for data augmentation. We pre-train the detector using the checkpoint released by InternImage~\cite{ii_model} and finetune it with the RDD2022 training data and the synthesized data from the generator until convergence.

For all the experiments, we follow the default experimental settings and parameters of the InternImage~\cite{ii} and CycleGAN~\cite{cyclegan} unless otherwise specified. We put all the other implementation details in the Appendix.

\section{Experimental result}
\label{sec:exp_result}

We first conduct an ablation study showing the efficacy of each component in \ours and summarize our experimental results in Table~\ref{ablation}, where we refer to each row by the experiment ID $E_1\sim E_5$. Except the column of parameter number (reported in million parameters), all the other numbers are reported in the F1 values (\%) on the RDD2022 testing set evaluated by the CRDDC 2022 challenge server. $E_1$ shows the performance of the SOTA method Faster Swin~\cite{sota_baseline}, but $E_5$ shows the final performance of our proposed \ours. $E_2 \sim E_5$ show that: (1) \ours outperforms the SOTA method; (2) each component in \ours contributes to the final performance; (3) using all of our proposed components reaches the best overall F1 value.

\begin{table}[t]
\centering
\resizebox{\columnwidth}{!}{%
\begin{tabular}{@{}c@{}c@{}c@{\hspace{1mm}}c@{\hspace{1mm}}c@{\hspace{2mm}}c@{\hspace{2mm}}c@{}}
\toprule
experiment ID &method &$\psi$ &$\cL_{\text{FID}}$ &$\cL_h$ &\# parameters (M)$\downarrow$ &overall (6 countries)\\
\midrule
$E_1$ &Faster Swin &\multicolumn{3}{c}{N/A} &253 &59.20 (+0)\\
\midrule
$E_2$ &\multirow{4}{*}{\ours} &\ccross &\ccross &\ccross &\textbf{49} &61.47 (+2.27)\\
$E_3$ & &$\ccheck$ &\ccross &\ccross  &\textbf{49} &62.52 (+3.32)\\
$E_4$ & &$\ccheck$ &$\ccheck$ &\ccross &\textbf{49} &62.94 (+3.74)\\
$E_5$ & &$\ccheck$ &$\ccheck$ &$\ccheck$ &\textbf{49} &\textbf{63.13 (+3.93)}\\
\bottomrule
\end{tabular}}
\caption{The ablation study comparing \ours with the top-performing method Faster Swin~\cite{sota_baseline} in the CRDDC 2022 challenge~\cite{crddc}. The numbers (except the parm. column) are the F1 values (\%) along with the gain over the Faster Swin on the RDD2022 testing set evaluated by the challenge server. Despite using less than 20\% of Faster Swin's parameters, \ours outperforms it on the average 6-country performance. Notations: $\psi$: the image generative model; $\cL_\text{FID}$: the Fréchet Inception Distance loss; $\cL_h$: the hard example synthesis loss.
}
\label{ablation}
\vspace{-.9em}
\end{table}

The fact that \ours only use $<$20\% of the number of parameters at the testing time compared with the Faster Swin shows that the performance improvement of \ours over the Faster Swin definitely does not come from using more parameters. Instead, the design choices of \ours, including using the InternImage-T~\cite{ii} backbone and the generative model, and the introduction of $\cL_\text{FID}$, and jointly training the generative model and detector with $\cL_h$ all contribute to the final performance of \ours. Comparing $E_2$ and $E_1$, we show that the usage of the InternImage-T backbone in \ours already improves the performance over the Faster Swin, even under fewer number of parameters. The ablation study of \ours presented in $E_2\sim E_5$ shows that the usage of the generative model $\psi$, training $\psi$ with the proposed Fréchet Inception Distance loss $\cL_\text{FID}$, and jointly training $\psi$ and the detector with $\cL_h$ further enhance the performance of \ours in the overall F1 value. $E_5$ shows that using all the aforesaid components reaches the best overall F1 value across the 6 countries. In the Appendix, we also show the qualitative ablation study of using $\cL_\text{FID}$.

Since the CRDDC 2022 challenge server also evaluates the country-wise performance of the four countries (India, Japan, Norway, and USA), we compare the country-wise performance of \ours versus that of the Faster Swin and report the result in Table~\ref{result}, where \ours outperforms the Faster Swin in 3 out of 4 country-wise performance and the mean of these 4 country-wise performance. The fact that $E_5$ is better than $E_1$ in most cases shows that \ours outperforms the SOTA method in country-wise and overall performance, regardless of whether we include the Czech Republic and China testing data for evaluation or not. Please notice that this evaluation is done on the held-out testing set (which is not publicly accessible) on the CRDDC 2022 challenge server, so unfortunately it is impossible for us to diagnose the failure cases from the testing data to understand why \ours does not outperform the Faster Swin in the USA testing data. However, we argue that in the USA testing data, \ours still performs competitively to the SOTA method Faster Swin (0.47 F1 score difference) while using only $<$20\% of the number of parameters. Furthermore, \ours outperforms the Faster Swin by much larger margins 1.31, 12.61, 3.38, 3.93 F1 scores in Japan, Norway, 4-country average, and 6-country average.

In addition, $E_5$ versus $E_4$ in Table 1 shows the efficacy of jointly training $\psi$ and $D$ using our proposed $\cL_h$, which supports that the data augmentation pipeline (\ie, $\psi$) used for object detection is better informed by the information from the detection objective instead of just being trained separately from the detector (\ie, $D$) in advance, as is typically done in the prior works of road defect detection. Although we only use a simple hard example mining loss $\cL_h$ for jointly training $\psi$ and $D$, the joint training strategy has the potential to incorporate more advanced design to train $\psi$ from the feedback given by $D$ (\eg, explicitly utilizing the confidence and accuracy of each detection output by $D$ to update the defect class frequency of $\psi$'s image synthesis strategy), which is part of our planned future work.

\section{Conclusion}
We propose \ours, a novel method for road defect detection that jointly trains a generative model and detector with dual discriminators and a CLIP-based Fréchet inception distance loss to enhance synthesized image quality and detector performance. Under the condition of not using the ensemble-based techniques or test-time augmentation to increase the test-time computational overhead, \ours outperforms the state-of-the-art methods on the RDD2022 benchmark across most country-wise and overall performance across six countries, while using less than 20\% of test-time parameters compared to baselines—making it well-suited for resource-constrained edge devices. Our ablations show that joint training the generative model and detector, improved image synthesis, and data augmentation boost the defect detection performance. Moreover, the dual discriminator design and CLIP-based Fréchet inception distance loss in \ours can serve as general enhancements for detection tasks beyond road defects.

{
    \small

\begin{thebibliography}{26}
\providecommand{\natexlab}[1]{#1}
\providecommand{\url}[1]{\texttt{#1}}
\expandafter\ifx\csname urlstyle\endcsname\relax
  \providecommand{\doi}[1]{doi: #1}\else
  \providecommand{\doi}{doi: \begingroup \urlstyle{rm}\Url}\fi

\bibitem[ii_()]{ii_model}
Internimage pre-trained model.
\newblock \url{https://tinyurl.com/2p825mt6}.

\bibitem[mot({\natexlab{a}})]{motive1}
News from the {Federal}.
\newblock \url{https://tinyurl.com/r39bdj8a}, {\natexlab{a}}.

\bibitem[mot({\natexlab{b}})]{motive2}
{AAA Newsroom}.
\newblock \url{https://tinyurl.com/ys9mn8b4}, {\natexlab{b}}.

\bibitem[pyt()]{pytorch}
{PyTorch}.
\newblock \url{https://pytorch.org/}.

\bibitem[Arya et~al.(2022{\natexlab{a}})Arya, Maeda, Ghosh, Toshniwal, Omata, Kashiyama, and Sekimoto]{crddc}
D. Arya, H. Maeda, S.~K. Ghosh, D. Toshniwal, H. Omata, T. Kashiyama, and Y. Sekimoto.
\newblock Crowdsensing-based road damage detection challenge {(CRDDC-2022)}.
\newblock \emph{arXiv preprint arXiv:2211.11362}, 2022{\natexlab{a}}.

\bibitem[Arya et~al.(2022{\natexlab{b}})Arya, Maeda, Ghosh, Toshniwal, and Sekimoto]{rdd}
D. Arya, H. Maeda, S.~K. Ghosh, D. Toshniwal, and Y. Sekimoto.
\newblock {RDD2022}: A multi-national image dataset for automatic road damage detection.
\newblock \emph{arXiv preprint arXiv:2209.08538}, 2022{\natexlab{b}}.

\bibitem[Bhavsar et~al.(2022)Bhavsar, Alfarrarjeh, Baranwal, and Kim]{sota_bhavsar}
M. Bhavsar, A. Alfarrarjeh, U. Baranwal, and S.~H. Kim.
\newblock Country-specific ensemble learning: A deep learning approach for road damage detection.
\newblock In \emph{IEEE International Conference on Big Data (Big Data)}, 2022.

\bibitem[Chakraborty et~al.(2024)Chakraborty, S, Naik, Panja, and Manvitha]{gan_survey}
T. Chakraborty, U.~R.~K S, S.~M. Naik, M. Panja, and B. Manvitha.
\newblock Ten years of generative adversarial nets ({GANs}): A survey of the state-of-the-art.
\newblock \emph{Machine Learning: Science and Technology}, 5\penalty0 (1), 2024.

\bibitem[Ding et~al.(2022)Ding, Zhao, Zhu, Du, Zhu, Yu, Li, and Wang]{sota_baseline}
W. Ding, X. Zhao, B. Zhu, Y. Du, G. Zhu, T. Yu, L. Li, and J. Wang.
\newblock An ensemble of one-stage and two-stage detectors approach for road damage detection.
\newblock In \emph{IEEE International Conference on Big Data (Big Data)}, 2022.

\bibitem[Everingham and Zisserman(2010)]{pascal}
Van Gool L. Williams C. K. I. Winn~J. Everingham, M. and A. Zisserman.
\newblock The {PASCAL} visual object classes ({VOC}) challenge.
\newblock \emph{IJCV}, 2010.

\bibitem[Heusel et~al.(2017)Heusel, Ramsauer, Unterthiner, Nessler, and Hochreiter]{fid}
M. Heusel, H. Ramsauer, T. Unterthiner, B. Nessler, and S. Hochreiter.
\newblock {GANs} trained by a two time-scale update rule converge to a local nash equilibrium.
\newblock In \emph{NIPS}, 2017.

\bibitem[Jeong and Kim(2022)]{sota_jeong}
D. Jeong and J. Kim.
\newblock Road damage detection using yolo with image tiling about multi-source images.
\newblock In \emph{IEEE International Conference on Big Data (Big Data)}, 2022.

\bibitem[Kynkäänniemi et~al.(2023)Kynkäänniemi, Karras, Aittala, Aila, and Lehtinen]{fid_clip}
T. Kynkäänniemi, T. Karras, M. Aittala, T. Aila, and J. Lehtinen.
\newblock The role of imagenet classes in {Fréchet} inception distance.
\newblock In \emph{ICLR}, 2023.

\bibitem[Lin et~al.(2014)Lin, Maire, Belongie, Bourdev, Girshick, Hays, Perona, Ramanan, Zitnick, and Dollár]{coco}
T.-Y. Lin, M. Maire, S. Belongie, L. Bourdev, R. Girshick, J. Hays, P. Perona, D. Ramanan, C.~L. Zitnick, and P. Dollár.
\newblock Microsoft {COCO}: Common objects in context.
\newblock In \emph{ECCV}, 2014.

\bibitem[Maeda et~al.(2020)Maeda, Kashiyama, Sekimoto, Seto, and Omata]{maeda}
H. Maeda, T. Kashiyama, Y. Sekimoto, T. Seto, and H. Omata.
\newblock Generative adversarial network for road damage detection.
\newblock In \emph{Computer-Aided Civil and Infrastructure Engineering}, 2020.

\bibitem[Mathiasen and Hvilshøj(2021)]{fid_loss}
A. Mathiasen and F. Hvilshøj.
\newblock Backpropagating through fréchet inception distance.
\newblock \emph{arXiv preprint arXiv:2009.14075}, 2021.

\bibitem[Okran et~al.(2022)Okran, Abdel-Nasser, Rashwan, and Puig]{sota_okran}
A.~M. Okran, M. Abdel-Nasser, H.~A. Rashwan, and D. Puig.
\newblock Effective deep learning-based ensemble model for road crack detection.
\newblock In \emph{IEEE International Conference on Big Data (Big Data)}, 2022.

\bibitem[Peng(2022)]{iskd}
K.-C. Peng.
\newblock Iterative self knowledge distillation — from pothole classification to fine-grained and covid recognition.
\newblock In \emph{ICASSP}, 2022.

\bibitem[Radford et~al.(2021)Radford, Kim, Hallacy, Ramesh, Goh, Agarwal, Sastry, Askell, Mishkin, Clark, Krueger, and Sutskever]{clip}
A. Radford, J.~W. Kim, C. Hallacy, A. Ramesh, G. Goh, S. Agarwal, G. Sastry, A. Askell, P. Mishkin, J. Clark, G. Krueger, and I. Sutskever.
\newblock Learning transferable visual models from natural language supervision.
\newblock In \emph{ICML}, 2021.

\bibitem[Russakovsky et~al.(2015)Russakovsky, Deng, Su, Krause, Satheesh, Ma, Huang, Karpathy, Khosla, Bernstein, et~al.]{imagenet}
O. Russakovsky, J. Deng, H. Su, J. Krause, S. Satheesh, S. Ma, Z. Huang, A. Karpathy, A. Khosla, M. Bernstein, et~al.
\newblock {ImageNet} large scale visual recognition challenge.
\newblock \emph{IJCV}, 115\penalty0 (3):\penalty0 211--252, 2015.

\bibitem[Saha and Sekimoto(2022)]{sota_saha}
P.~K. Saha and Y. Sekimoto.
\newblock Road damage detection for multiple countries.
\newblock In \emph{IEEE International Conference on Big Data (Big Data)}, 2022.

\bibitem[Wang et~al.(2022)Wang, Tang, Liao, He, Feng, Jiao, Su, and Yuan]{sota_wang}
S. Wang, Y. Tang, X. Liao, J. He, H. Feng, H. Jiao, X. Su, and Q. Yuan.
\newblock An ensemble learning approach with multi-depth attention mechanism for road damage detection.
\newblock In \emph{IEEE International Conference on Big Data (Big Data)}, 2022.

\bibitem[Wang et~al.(2023)Wang, Dai, Chen, Huang, Li, Zhu, Hu, Lu, Lu, Li, Wang, and Qiao]{ii}
W. Wang, J. Dai, Z. Chen, Z. Huang, Z. Li, X. Zhu, X. Hu, T. Lu, L. Lu, H. Li, X. Wang, and Y. Qiao.
\newblock {InternImage}: Exploring large-scale vision foundation models with deformable convolutions.
\newblock In \emph{CVPR}, 2023.

\bibitem[Wu et~al.(2022)Wu, Liao, Zhang, Wang, Bai, Cheng, and Liu]{yolop}
D. Wu, M. Liao, W. Zhang, X. Wang, X. Bai, W. Cheng, and W. Liu.
\newblock {YOLOP}: You only look once for panoptic driving perception.
\newblock \emph{Machine Intelligence Research}, 2022.

\bibitem[Zhong et~al.(2023)Zhong, Huyan, Zhang, Cheng, Zhang, Tong, Jiang, and Huang]{zhong}
J. Zhong, J. Huyan, W. Zhang, H. Cheng, J. Zhang, Z. Tong, X. Jiang, and B. Huang.
\newblock A deeper generative adversarial network for grooved cement concrete pavement crack detection.
\newblock In \emph{Engineering Applications of Artificial Intelligence}, 2023.

\bibitem[Zhu et~al.(2017)Zhu, Park, Isola, and Efros]{cyclegan}
J.-Y. Zhu, T. Park, P. Isola, and A.~A. Efros.
\newblock Unpaired image-to-image translation using cycle-consistent adversarial networks.
\newblock In \emph{ICCV}, 2017.

\end{thebibliography}

}

\clearpage
\setcounter{page}{1}
\section*{\Large Appendix}

\setcounter{section}{0}
\renewcommand{\theHsection}{A\arabic{section}}
\renewcommand{\thesection}{A\arabic{section}}
\renewcommand{\thetable}{A\arabic{table}}
\setcounter{table}{0}
\setcounter{figure}{0}
\renewcommand{\thetable}{A\arabic{table}}
\renewcommand\thefigure{A\arabic{figure}}
\renewcommand{\theHtable}{A.Tab.\arabic{table}}
\renewcommand{\theHfigure}{A.Abb.\arabic{figure}}
\renewcommand\theequation{A\arabic{equation}}
\renewcommand{\theHequation}{A.Abb.\arabic{equation}}

\noindent The appendix is organized as follows:
\begin{itemize}[topsep=0pt, leftmargin=16pt]
\item In Sec.~\ref{sec:rdd_stats}, we provide the statistics of the RDD 2022 dataset~\cite{rdd} and the details of the evaluation protocol.
\item In Sec.~\ref{sec:dad_ds}, we provide the details of drivable area detection and defect synthesis.
\item In Sec.~\ref{sec:implementation}, we provide the implementation details of \ours, including the hyperparameter settings and hardware specification.
\item In Sec.~\ref{sec:qualitative_ablation}, we provide the qualitative ablation study of using our proposed Fréchet Inception Distance loss $\cL_\text{FID}$.
\end{itemize}

\section{The statistics of the RDD 2022 dataset~\cite{rdd} and the details of the evaluation protocol}
\label{sec:rdd_stats}

\begin{figure*}
\centering
\includegraphics[width=.8\linewidth]{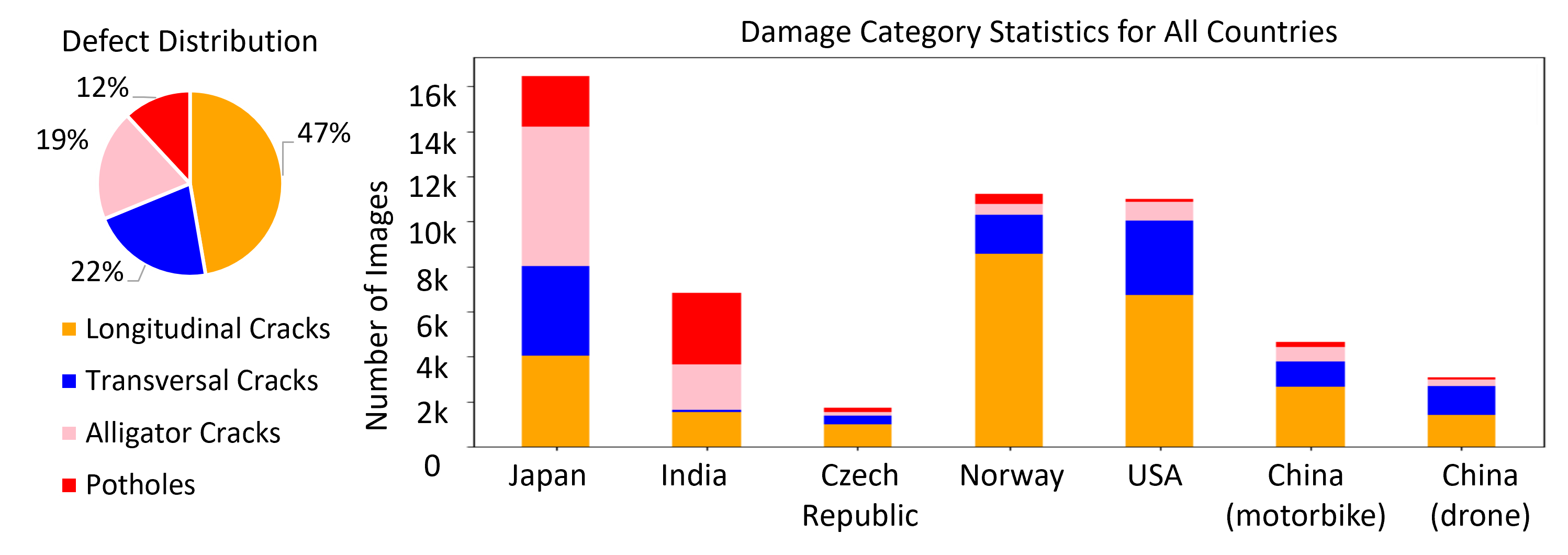}
\caption{The statistics of the RDD2022 dataset~\cite{rdd}.}
\label{rdd_stats}
\end{figure*}

The RDD2022 dataset~\cite{rdd} includes the road images from 6 countries and annotations for at least 4 types of defects, as shown in Fig.~\ref{rdd_stats}. In terms of the evaluation protocol, we use the official training split and evaluation protocol of the CRDDC 2022 challenge~\cite{crddc}. Although the RDD2022 dataset includes the images from 6 countries, the evaluation server only provides country-wise evaluation for Japan, India, Norway, and USA, and overall evaluation for all 6 countries on the holdout testing set. For simplicity, we train one model using the entire training set and evaluate it via the evaluation server on the holdout country-specific and overall testing sets. This protocol is motivated by the need for a system that is less computationally demanding without the burden of switching between multiple models for different scenarios. We report F1 values as the evaluation metric.

\section{The details of drivable area detection and defect synthesis}
\label{sec:dad_ds}
\begin{figure*}
\centering
\includegraphics[width=\linewidth]{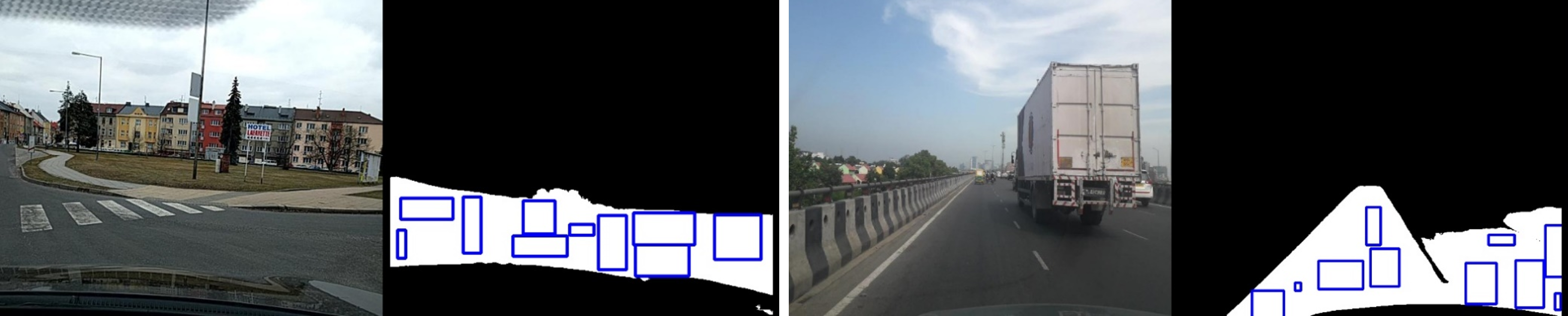}
\caption{The examples of drivable area detection by YOLOP~\cite{yolop}. The bounding boxes of the input mask of the generator of \ours are created in the drivable areas, \eg, the blue boxes in the examples.}
\label{drivable}
\end{figure*}

To make sure that we do not synthesize the defects outside of the road areas, we use YOLOP~\cite{yolop} to perform drivable area detection and enforce the bounding boxes in the input mask of the generator to be inside the detected drivable areas, as illustrated in Fig.~\ref{drivable}. These bounding boxes in the mask are created at random location within the detected drivable areas with a random aspect ratio and a random size such that any two bounding boxes do not overlap. We assign a random defect type out of the four defect classes for each bounding box in the input mask.

\section{The implementation details of \ours}
\label{sec:implementation}
For the generator and detector in \ours, We use the initial learning rate 1e-4, weight decay 1e-4, batch size 16, and the Adam optimizer to train them until convergence. We train the generator and detector of \ours on four 48GB NVIDIA A40 GPUs. We empirically set $w_h=1$ and $w_\text{FID}=0.1$ such that the ranges of $\cL_h$ and $\cL_\text{FID}$ are comparable to other loss terms. We choose the confidence threshold of the detector via the validation data sampled from the training set. For all the other parameters, since we implement \ours using PyTorch~\cite{pytorch}, we follow its default setting unless otherwise specified.

\section{The qualitative ablation study of using our proposed Fréchet Inception Distance loss}
\label{sec:qualitative_ablation}
\begin{figure*}
\centering
\includegraphics[width=.7\linewidth]{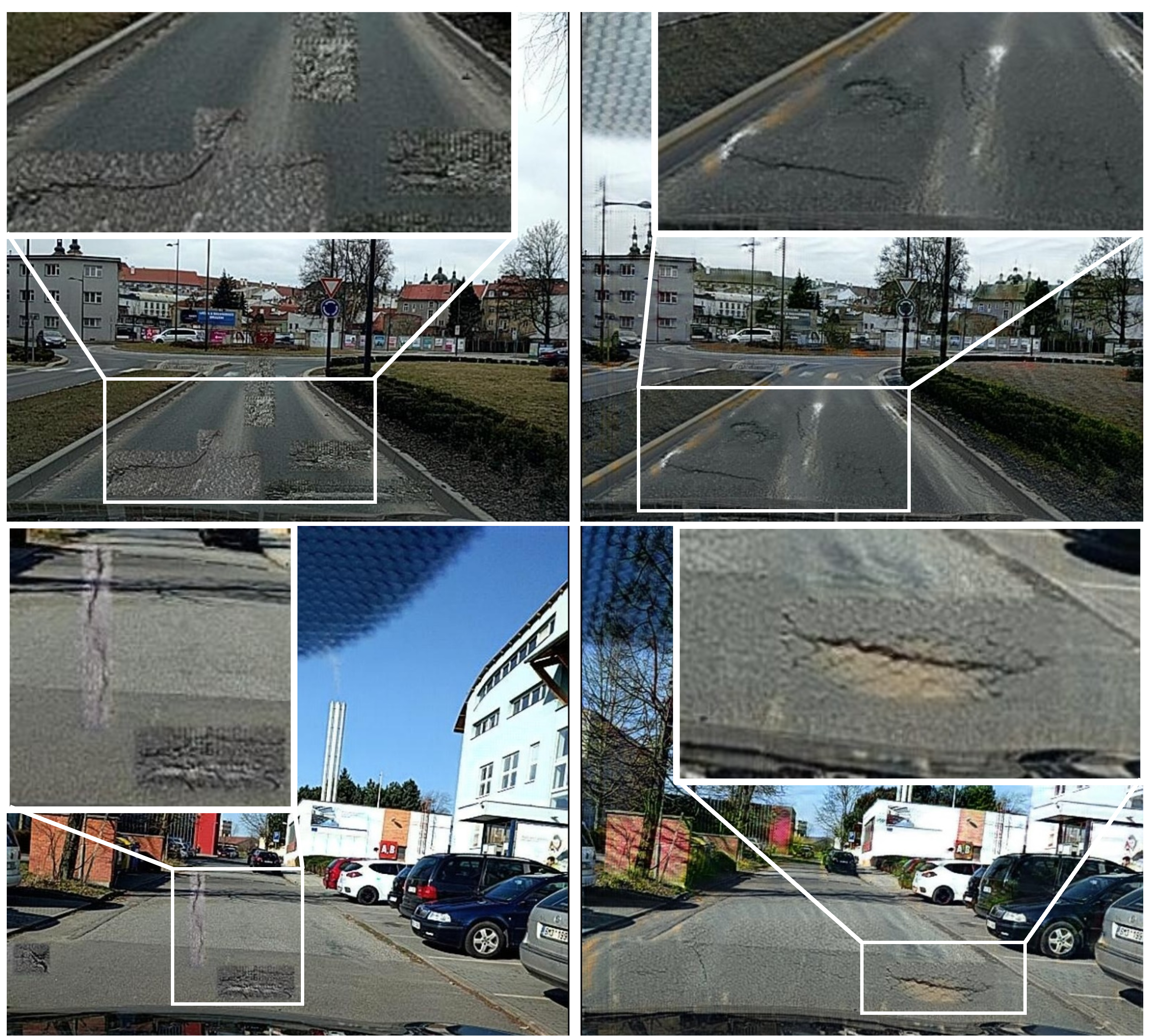}
\\
(a) w/o $\cL_\text{FID}$ \hspace{1.7in} (b) w/ $\cL_\text{FID}$
\vspace{.2em}
\caption{The qualitative ablation study of using our proposed Fréchet Inception Distance loss $\cL_\text{FID}$.}
\label{fid}
\end{figure*}

To gain more intuitive insights about the impact of using our proposed Fréchet Inception Distance loss $\cL_\text{FID}$, we show the qualitative ablation study of using $\cL_\text{FID}$ or not in Fig.~\ref{fid}, where the areas of the synthesized defects are enlarged for visibility. Fig.~\ref{fid} (a) shows that even if we already use the dual discriminator design when training the generative model, without using $\cL_\text{FID}$, we can still see some obvious artifacts around the boundaries of the added defects which cannot be fully eliminated by training the generator with the global discriminator. In contrast, Fig.~\ref{fid} (b) shows that the artifacts around the boundaries of the added defects are less obvious by training the generator with $\cL_\text{FID}$, and hence the overall quality of the synthesized images is improved. Quantitatively, we evaluate the Fréchet Inception Distance (FID) between the synthesized images and the real training images and find that training with $\cL_\text{FID}$ (Fig.~\ref{fid} (b)) improves the FID (lower is better) from 37.04 to 32.68 compared with training without $\cL_\text{FID}$ (Fig.~\ref{fid} (a)). Considering the aforesaid quantitative FID comparison, Fig.~\ref{fid}, and $E_3$ versus $E_4$ in Table~\ref{ablation}, we show that improving the quality of the synthesized images in the data augmentation process can improve the performance of the defect detector. Although we use the CycleGAN~\cite{cyclegan} as the backbone of the generative model in \ours for road defect detection, we expect that the application of $\cL_\text{FID}$ and dual discriminator design is agnostic to the backbone of the generative models and the object of interest in the detection task as long as the gradient associated with $\cL_\text{FID}$ can be backpropagated through the backbone of the generative model. Given that FID has become a standardized evaluation metric for assessing image quality in image synthesis literature~\cite{gan_survey}, it makes sense to directly minimize FID as one of the optimization objectives to enhance the image synthesis quality, and hence improve the performance of defect detection using the augmented data with better quality, as what we design for \ours. 

\end{document}